\documentclass[runningheads]{llncs}

\usepackage[utf8]{inputenc} \usepackage[T1]{fontenc} \usepackage{graphicx} 
\usepackage{hyperref} %\usepackage{caption} %\usepackage{subcaption}

\begin{document} \title{Toward an Integrated Cognitive--Ergonomic Architecture for Human--Machine Interaction: Combining Cognitive Models with Human Factors Ergonomics} \author{Antoine LENAT\inst{1,2}, Olivier CHEMINAT\inst{2}, Damien CHABLAT\inst{1}, Camilo CHARRON\inst{1,3}}
\authorrunning{A. LENAT et al.}
\institute{
$^1$Nantes Université, École Centrale Nantes, CNRS, LS2N, UMR 6004 \\
44000 Nantes, France\\
$^2$ Université Rennes 2, Rennes, France\\
$^3$ CETIM, 74 Route de la Jonelière, 44300 Nantes \\
\email{\{Antoine.Lenat, Camilo.Charron, Damien.Chablat\}@ls2n.fr}}
\institute{
$^1$Nantes Université, École Centrale Nantes, CNRS, LS2N, UMR 6004 \\
44000 Nantes, France\\
$^2$ CETIM, 74 Route de la Jonelière, 44300 Nantes \\
$^3$ Université Rennes 2, Rennes, France\\
\email{\{Antoine.Lenat, Damien.Chablat, Camilo.Charron\}@ls2n.fr}, Olivier.Cheminat@cetim.fr}

\maketitle

\begin{abstract} 
This paper presents an integrated approach to modeling human competencies by combining the theoretical foundations of cognitive architectures with principles from Human Factors Ergonomics (HFE). Through a comparative analysis of established cognitive models—SOAR, ACT-R, LIDA, and COCOM—we synthesize a tailored architecture designed to address the complexities of human--machine interaction (HMI) in dynamic environments. By contextualizing this model within ergonomic frameworks, we elucidate the mechanisms underlying decision-making, skill acquisition, and adaptive behavior, bridging the gap between cognitive theory and applied system design.
Our framework is empirically grounded in industrial robotics applications, where operator expertise, normative knowledge, and real-time feedback loops are critical. The proposed architecture not only enhances the cognitive alignment of HMI systems but also provides a scalable methodology for designing intelligent, human-centered interfaces in high-stakes environments. This work advances both the theoretical understanding of human competencies and the practical implementation of adaptive, ergonomically optimized systems.

\keywords{Cognitive architectures \and Human Factors Ergonomics \and Human--Machine Interaction \and Competence modeling} \end{abstract}
%%%%%%%%%%%%%%%%%%%%%%%%%%%%%%%%%%%%%%%%%%%%%%
\section{Introduction} 
%%%%%%%%%%%%%%%%%%%%%%%%%%%%%%%%%%%%%%%%%%%%%%
Human--machine interaction (HMI) in safety-critical and knowledge-intensive domains requires models capable of accounting for cognition, action, context, and adaptation. Cognitive architectures have long aimed to formalize the mechanisms underlying human reasoning and learning, while Human Factors Ergonomics (HFE) focuses on the conditions that shape human activity in real work situations \cite{leplat_regards_1997,hoc_analyse_1999}. Despite their shared objectives, these traditions have largely evolved in parallel.

Recent studies in the field of welding operations have highlighted the importance of fast adaptation and domain-specific knowledge in guiding the set of parameters \cite{lenat_improving_2024,lenat_novel_2025}. A heuristic map has been used to represent all variables impacting the quality of the weld bead, identified through literature review and participant observation. This map is used in interviews with engineers to gather qualitative information, providing a diagnostic tool and a foundation for further investigation into the nature of welding expertise \cite{vermersch_entretien_2019,antoine_saisir_2017}.

This article builds on prior work in cognitive modeling and ergonomics to propose an integrated cognitive--ergonomic architecture. By combining symbolic and subsymbolic cognitive models with ergonomic concepts such as activity regulation, situation awareness, and competence, we aim to bridge the gap between internal cognitive mechanisms and observable human activity in complex systems.

The remainder of this paper first presents the theoretical foundations of cognition and activity, then reviews major cognitive architectures, before introducing an integrated cognitive-ergonomic architecture and illustrating its relevance through applications in industrial robotics and critical interface design, followed by a discussion and perspectives for future work.

%%%%%%%%%%%%%%%%%%%%%%%%%%%%%%%%%%%%%%%
\section{Theoretical Foundations} 
\subsection{Cognition, Activity, and Competence} 
\begin{figure}[!ht]
    \centering
    \includegraphics[width=0.8\textwidth]{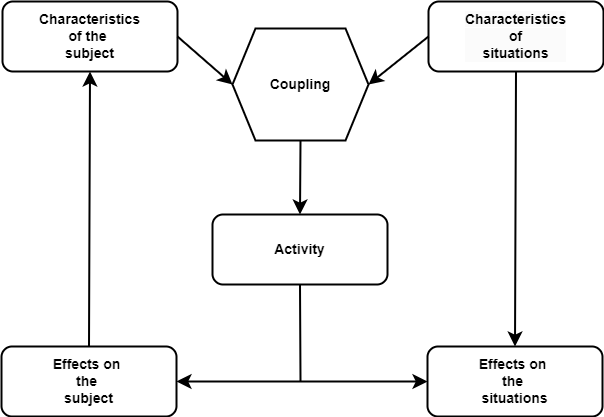}
    \caption{Double regulation of activity linking task constraints and internal regulation processes.}
    \label{fig:double_regulation}
\end{figure}

%%%%%%%%%%%%%%%%%%%%%%%%%%%%%%%%%%%%%%%
From an ergonomic standpoint, cognition is inseparable from activity. Human performance emerges from the dynamic regulation between internal states, task constraints, and environmental resources \cite{leplat_regards_1997,amalberti_maitrise_2001}. This principle of double regulation, as illustrated in Figure~\ref{fig:double_regulation}, constitutes a cornerstone of francophone ergonomics, linking task constraints and internal regulation processes. Competence is not a static attribute but an organized and evolving structure that enables adaptive behavior across situations \cite{vidal-gomel_competences_2016,coulet_organization_2019}.

Drawing on ethnographic data, a welding scheme applicable to both GTAW and GMAW processes has been developed. A superordinate scheme was also modeled to describe the operator’s training process, in which optimal task settings are determined through simplified situations that reduce cognitive load. These two processes differ in their subschemes, particularly in the gesture vector (i.e., action rules), where the manipulation of filler metal varies depending on the welding technique.

The next step involves experimental validation with welders to assess the robustness of the welding scheme, the training scheme (as a superordinate structure), and the feedback scheme (a subscheme of the welding process). These tests will help identify the specific classes of situations in which each scheme applies and determine whether the organization of activity remains consistent across task repetitions.

The model may be challenged if activity organization proves highly variable—i.e., if operator behavior changes significantly with each repetition of the same task. It is therefore essential to identify the specific classes of situations where the scheme remains valid. Additionally, the description of subschemes may be incomplete due to the wide range of possible scenarios. The model relies on a form of pattern matching between current and past situations; if an operator encounters a situation that cannot be approximated by a known one, the model would require substantial adaptation, which may not be feasible, particularly in robotic applications.
%%%%%%%%%%%%%%%%%%%%%%%%%%%%%%%%%%%%%%%
\subsection{Memory, Knowledge, and Cognitive Resources} 
%%%%%%%%%%%%%%%%%%%%%%%%%%%%%%%%%%%%%%%
Research in cognitive psychology highlights the limits of working memory and the role of long-term knowledge structures in skilled performance \cite{barrouillet_memoire_2022,gobet_psychologie_2011}. These mechanisms are essential for understanding error, learning, and expertise, and they form a natural bridge between cognitive architectures and ergonomic analyses of task complexity.

In the context of welding operations, the importance of fast adaptation and domain-specific knowledge in guiding the set of parameters has been emphasized. A heuristic map has been developed to represent all variables impacting the quality of the weld bead, identified through literature review and participant observation. This map is used in interviews with engineers to gather qualitative information, providing a diagnostic tool and a foundation for further investigation into the nature of welding expertise \cite{lenat_improving_2024,lenat_novel_2025}.

The integration of robots in industrial settings is increasingly common, addressing both the deficit in skilled workers in manual professions and the health risks associated with certain tasks. These health risks include posture-related issues (musculoskeletal disorders), exposure to harmful substances (toxic fumes, radiation), or repetitive motions. Welding is a prime example, with a shortage of skilled operators in France to meet industry needs. This evolution responds to the need for systems that can adapt to complex and variable environments, where human skill and cognitive flexibility are crucial.
%%%%%%%%%%%%%%%%%%%%%%%%%%%%%%%%%%%%%%%
\section{Cognitive Architectures for Human Modeling} 
%%%%%%%%%%%%%%%%%%%%%%%%%%%%%%%%%%%%%%%
This section reviews major cognitive architectures used for modeling human cognition and action, and discusses their application in the context of human skill modeling, particularly in welding operations.
\subsection{SOAR}
SOAR is a symbolic cognitive architecture based on problem spaces and production rules. Learning occurs through chunking mechanisms that compile experience into procedural knowledge \cite{laird_soar_2012,laird_standard_2017}. SOAR has been widely used to model complex problem-solving tasks and has been applied in various domains, including robotics and human-machine interaction.
\subsection{ACT-R}
ACT-R integrates symbolic rules with subsymbolic activation processes governing memory retrieval and action timing \cite{chase_minds_1973}. Its modular organization makes it suitable for modeling perception, cognition, and motor interaction. ACT-R has been particularly useful in understanding how knowledge is structured and retrieved in skilled performance.
\subsection{LIDA}
The LIDA model emphasizes perception-action cycles and attention mechanisms inspired by the Global Workspace Theory \cite{rosenbloom_rethinking_2016}. It provides a biologically grounded approach to adaptive cognition, making it suitable for modeling tasks that require real-time regulation and feedback control.
\subsection{COCOM}
COCOM focuses on control modes governing human behavior in dynamic and time-critical environments \cite{hollnagel_context_1998,rasmussen_skills_1983}. It distinguishes between strategic, tactical, opportunistic, and scrambled control, and is particularly relevant for modeling human performance in complex, safety-critical domains.
%%%%%%%%%%%%%%%%%%%%%%%%%%%%%%%%%%%%%%%
\subsection{Application in Welding Operations}
%%%%%%%%%%%%%%%%%%%%%%%%%%%%%%%%%%%%%%%
As introduced in previous studies \cite{lenat_improving_2024,lenat_novel_2025}, our research builds on the scheme theory \cite{vergnaud_theory_2009}, developed within the conceptual fields framework, to investigate the competencies of manual operators. By integrating this theory with cognitive architecture models \cite{albus_rcs_2005,laird_soar_2012,laird_standard_2017}, we propose a methodology to characterize operator skills, using welding as a case study.

Our objective is first to analyze the cognitive abilities mobilized by manual operators and then to represent these abilities in a graphical model. This model aims to deconstruct the cognitive and gestural processes used by operators with the goal of informing robotic systems and advancing automation in complex manual tasks.

We formulate several key assumptions. We consider schemes and competencies as conceptually equivalent, with schemes serving as a framework for modeling skill. These schemes may be individual or shared among multiple operators. We hypothesize that they can be identified by confronting expert knowledge, such as engineer insights or literature, with the practical know-how from operators. We also acknowledge the presence of potentially false beliefs, reinforcing the need for controlled experimental settings. Finally, we assume that skill acquisition involves both implicit and explicit learning processes, which complicates its formalization and sharing.

To increase robotic robustness, we propose to model human skill, helping in the same way to improve the training of operators. Using a better strategy of robotization can facilitate communication among multiple contributors, such as roboticists, operators, and engineers.

\begin{figure}[!ht]
    \centering
    \includegraphics[width=\textwidth]{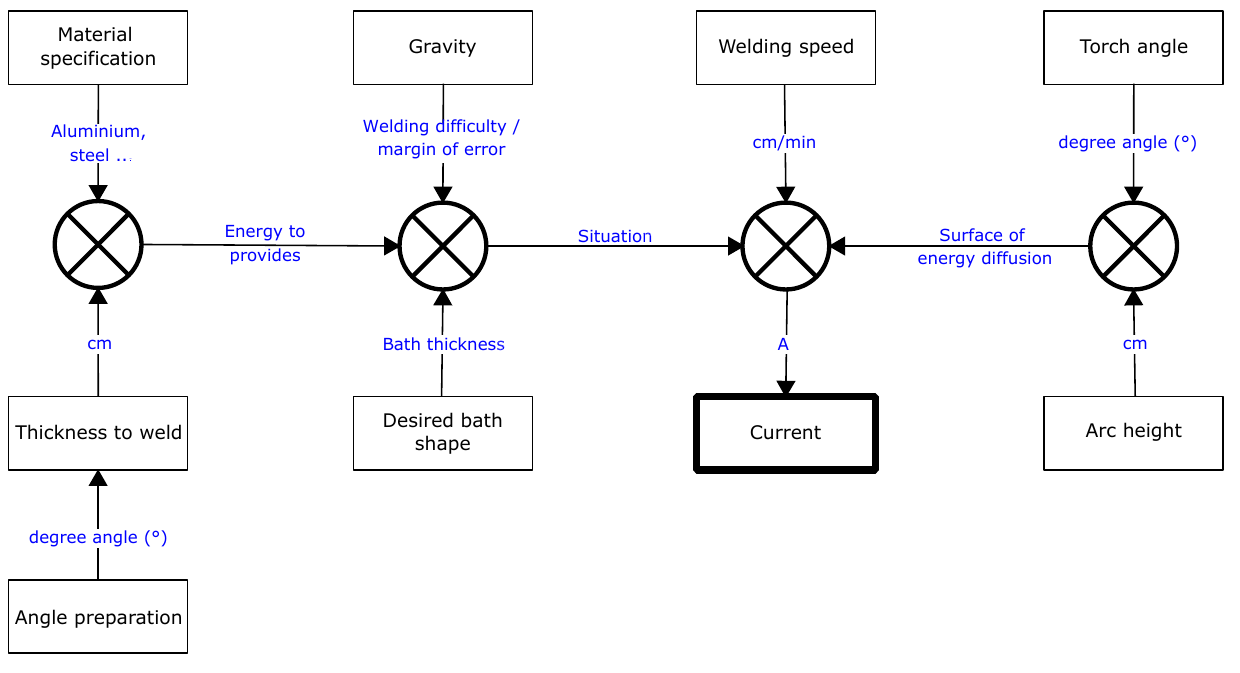}
    \caption{Flowchart of the current selection scheme illustrating the information flow used to define the current setting.}
    \label{fig:SchemeConsigne}
\end{figure}

The current selection scheme (Figure~\ref{fig:SchemeConsigne}) was modeled using a block diagram-like graph, which serves as a foundation for graph-based or probabilistic representations like Bayesian networks. These are particularly useful when dealing with numerous variables, both qualitative and quantitative. The goal is not to derive a precise equation for current settings, but to orient decisions toward optimal values, especially when results are non-compliant.

Focusing on an action as a subscheme allows for a deeper understanding of the variables involved, the flow of information, the tools required, and the knowledge mobilized. The chosen representation should capitalize on the scheme’s structure—whether through inference, goals, elementary tasks, or operational invariants.

Even though this article does not describe full robotization, it lays the groundwork for translating a cognitive model into operational automated systems.
%%%%%%%%%%%%%%%%%%%%%%%%%%%%%%%%%%%%%%%
\section{Human Factors Ergonomics Perspective}
\label{sec:human_factors_ergonomics}
%%%%%%%%%%%%%%%%%%%%%%%%%%%%%%%%%%%%%%%
Human Factors Ergonomics emphasizes the analysis of real activity rather than prescribed tasks. Variability in performance is interpreted as adaptive regulation rather than deviation from norms \cite{leplat_regards_1997,amalberti_maitrise_2001}. This perspective is particularly relevant in complex and dynamic work environments, where operators must continuously adapt to changing conditions and constraints.
%%%%%%%%%%%%%%%%%%%%%%%%%%%%%%%%%%%%%%%
\subsection{Operator Expertise and Normative Knowledge}
%%%%%%%%%%%%%%%%%%%%%%%%%%%%%%%%%%%%%%%
Operator expertise is also regulated through qualification standards \cite{iso_9606-1_qualification_2017} and safety protocols \cite{iec_62822-1_electric_2018}, reinforcing the role of normative knowledge. Thus, welding knowledge is deeply embedded in normative structures that guide execution, evaluation, and transmission—making standards a key entry point for analyzing skill and professional identity.
%%%%%%%%%%%%%%%%%%%%%%%%%%%%%%%%%%%%%%%
\subsection{Participant Observation}
%%%%%%%%%%%%%%%%%%%%%%%%%%%%%%%%%%%%%%%
To grasp the complexity of welding as a profession, two participant observations were conducted. The first involved working in an office among engineers for two years, and the second, during an initiation training program for one week. This immersive experience offers a general yet meaningful understanding of the welder’s environment, highlighting key ergonomic, sensory, and psychological dimensions such as posture, visibility, fatigue, and stress management.

The collected data, such as verbatim transcripts, sensory impressions, and reflective notes, were synthesized in the form of a logbook and experience feedback. This method supports a deeper engagement with the profession and enhances the quality of subsequent qualitative inquiries \cite{olivier_de_sardan_rigueur_2008}.

This welding training program is designed to provide a holistic overview of the welding profession, emphasizing the welder's critical role in the manufacturing process and their mastery of quality standards, particularly in ensuring weld bead conformity. The insights for researchers are the abilities to access embodied knowledge \cite{varela_embodied_1993} and develop empathy \cite{brugeron_analyzing_2023}, which are essential for conducting insightful interviews and conceptualizing the profession. The observation also facilitates a comparative understanding of welding techniques (GTAW and GMAW), their principles, and applications.
%%%%%%%%%%%%%%%%%%%%%%%%%%%%%%%%%%%%%%%
\subsection{A Multiphysics System Model}
\begin{figure}[!htbp]
    \centering
    \includegraphics[width=\textwidth]{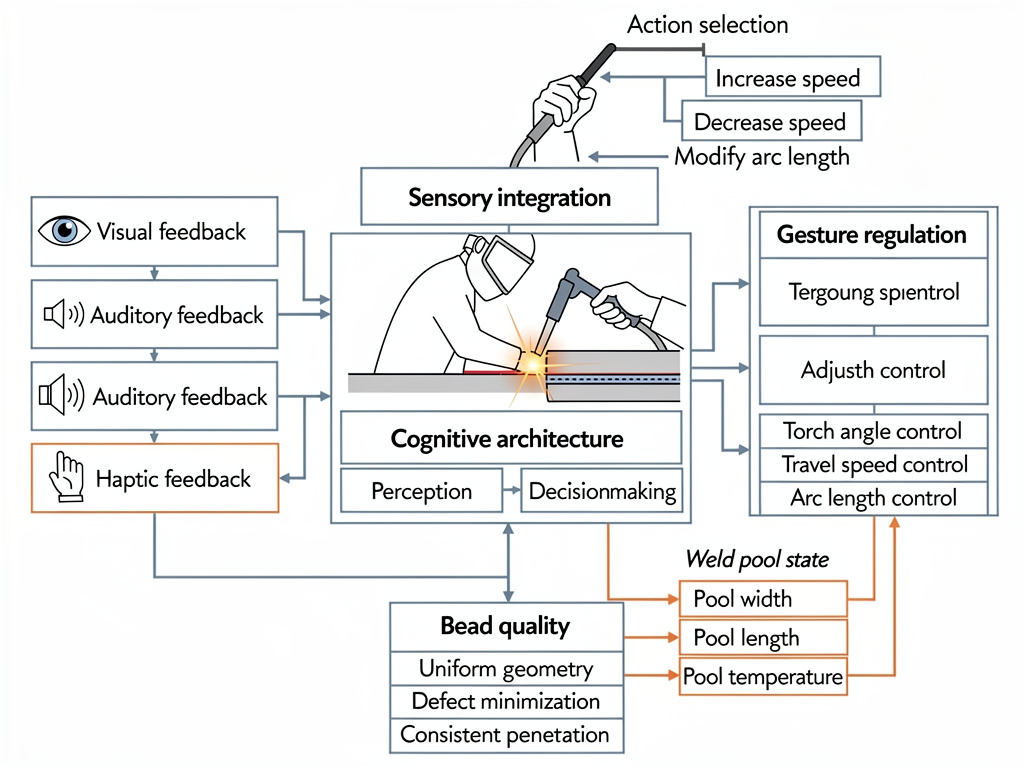}
    \caption{Block diagram of the welding process as a multiphysics system, illustrating sensory integration, gesture regulation, and cognitive architecture for real-time control and decision-making.}
    \label{fig:welding_system}
\end{figure}

%%%%%%%%%%%%%%%%%%%%%%%%%%%%%%%%%%%%%%%
The welding process can be conceptualized as a complex multiphysics system, where the operator's primary focus lies in the precise manipulation of the weld pool. To achieve a uniform bead, the operator relies on continuous sensory feedback and gesture regulation, such as adjusting the torch angle and travel speed. This dynamic interaction is modeled through a cognitive architecture, integrating real-time feedback control and decision-making processes. Figure~\ref{fig:welding_system} illustrates this conceptual framework, highlighting the interplay between sensory integration, gesture regulation, and cognitive decision-making in maintaining weld quality.

%%%%%%%%%%%%%%%%%%%%%%%%%%%%%%%%%%%%%%%
\section{Discussion and future work} 
%%%%%%%%%%%%%%%%%%%%%%%%%%%%%%%%%%%%%%%
In this paper, we have proposed an integrated cognitive-ergonomic architecture designed to bridge the gap between cognitive modeling and human factors ergonomics in human-machine interaction. By combining theoretical foundations from cognitive science with practical insights from ergonomics, we have demonstrated how this architecture can enhance the understanding of operator expertise, task regulation, and adaptive behavior in complex systems.

\begin{figure}[!ht]
    \centering
    \includegraphics[width=\textwidth]{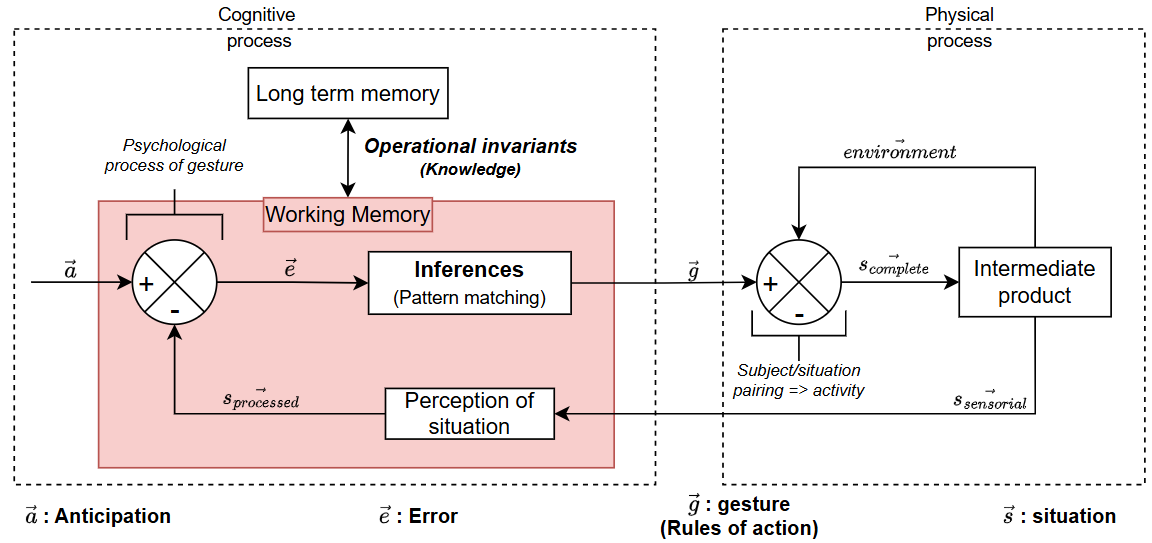}
    \caption{Skill model integrating Vergnaud's scheme and cognitive architectures - feedback scheme}    
    \label{fig:FeedbackScheme}
\end{figure}

Our exploration of cognitive architectures such as SOAR, ACT-R, LIDA, and COCOM, alongside empirical observations from welding operations, has underscored the importance of a holistic approach to human-machine systems. The proposed framework not only aligns with Vergnaud's scheme of activity but also integrates feedback mechanisms to support real-time decision-making and skill acquisition as shown in Figure~\ref{fig:FeedbackScheme}.

The application of this architecture in real-world scenarios, such as welding operations, highlights its potential to improve both operator performance and system design. By formalizing the interplay between normative knowledge and operator expertise, we provide a foundation for developing more intuitive and adaptive human-machine interfaces.

Looking ahead, future work will focus on validating this architecture in diverse industrial contexts and refining its computational models to accommodate a broader range of cognitive and ergonomic factors. Additionally, exploring the integration of machine learning techniques could further enhance the adaptability and predictive capabilities of the system.

Ultimately, this research contributes to the broader goal of creating human-centered technologies that are not only efficient but also aligned with the cognitive and ergonomic needs of their users.
%%%%%%%%%%%%%%%%%%%%%%%%%%%%%%%%%%%%%%%
\section*{Acknowledgments}
%%%%%%%%%%%%%%%%%%%%%%%%%%%%%%%%%%%%%%%
This work was supported by the French National Association for Research and Technology (ANRT) through the Industrial Conventions of Training through Research (CIFRE) program under grant number CIFRE n°2022/1562 for the funding of the first author.

\bibliographystyle{splncs04} 
\bibliography{bib-list-2025}

\end{document}